\documentclass{article}

\usepackage{microtype}
\usepackage{graphicx}
\usepackage{booktabs} % for professional tables

\usepackage[hidelinks]{hyperref}

\usepackage[nohyperref, accepted]{icml2024}

\usepackage{amsmath}
\usepackage{amssymb}
\usepackage{mathtools}
\usepackage{amsthm}

\usepackage{enumitem}
\usepackage{multicol}
\usepackage{multirow}
\usepackage{wrapfig}
\usepackage{subcaption}
\usepackage{nicefrac}
\usepackage{tikz}

\usepackage{amsmath, amsfonts, bm, tabstackengine}
\newcommand{\emptyargs}{({}\cdot{})}
\DeclareMathOperator*{\update}{\textsc{Up}}
\DeclareMathOperator*{\msg}{\textsc{Msg}}
\DeclareMathOperator*{\aggr}{\textsc{Aggr}}

\DeclareMathOperator*{\select}{\textsc{Sel}}
\DeclareMathOperator*{\connect}{\textsc{Con}}
\DeclareMathOperator*{\reduce}{\textsc{Red}}
\DeclareMathOperator*{\lift}{\textsc{Lift}}
\newcommand{\mlp}{\text{MLP}}
\newcommand{\gnn}{\text{GNN}}
\def\bigO{\mathcal{O}}

\def\eqref#1{equation~\ref{#1}}
\def\1{\bm{1}}

\def\vb{{\bm{b}}}

\def\vh{{\bm{h}}}

\def\vu{{\bm{u}}}
\def\vv{{\bm{v}}}

\def\vx{{\bm{x}}}
\def\vy{{\bm{y}}}

\def\mA{{\bm{A}}}

\def\mC{{\bm{C}}}
\def\mD{{\bm{D}}}

\def\mH{{\bm{H}}}
\def\mI{{\bm{I}}}

\def\mP{{\bm{P}}}
\def\mQ{{\bm{Q}}}
\def\mR{{\bm{R}}}
\def\mS{{\bm{S}}}

\def\mU{{\bm{U}}}
\def\mV{{\bm{V}}}
\def\mW{{\bm{W}}}
\def\mX{{\bm{X}}}
\def\mY{{\bm{Y}}}
\def\mZ{{\bm{Z}}}

\def\mPhi{{\bm{\Phi}}}

\DeclareMathAlphabet{\mathsfit}{\encodingdefault}{\sfdefault}{m}{sl}
\SetMathAlphabet{\mathsfit}{bold}{\encodingdefault}{\sfdefault}{bx}{n}

\def\gG{{\mathcal{G}}}

\def\sR{{\mathbb{R}}}

\usepackage[acronym, nohypertypes={acronym}]{glossaries}

\newacronym{method}{HiGP}{\textit{Hierarchical Graph Predictor}}
\newacronym{fr}{FR}{\textit{forecast reconciliation}}

\newacronym{stgnn}{STGNN}{spatiotemporal graph neural network}
\newacronym{tts}{TTS}{time-then-space}
\newacronym{mlp}{MLP}{multi-layer perceptron}
\newacronym{rnn}{RNN}{recurrent neural network}
\newacronym{gnn}{GNN}{graph neural network}
\newacronym{src}{SRC}{\textit{select, reduce, connect}}

\newglossaryentry{la}{name=Metr-LA,description=}
\newglossaryentry{bay}{name=PeMS-Bay,description=}
\newglossaryentry{cer}{name=CER-E,description=}
\newglossaryentry{air}{name=AQI,description=}
\newglossaryentry{pems3}{name=PEMS03,description=}
\newglossaryentry{pems4}{name=PEMS04,description=}
\newglossaryentry{pems7}{name=PEMS07,description=}
\newglossaryentry{pems8}{name=PEMS08,description=}

\newglossaryentry{ttsimp}{name=TTS-IMP,description=}
\newglossaryentry{ttsamp}{name=TTS-AMP,description=}
\newglossaryentry{tasimp}{name=T\&S-IMP,description=}
\newglossaryentry{tasamp}{name=T\&S-AMP,description=}

\theoremstyle{plain}

\theoremstyle{definition}

\theoremstyle{remark}

\usepackage[textsize=tiny]{todonotes}

\makeatletter
\newcommand{\pushright}[1]{\ifmeasuring@#1\else\omit\hfill$\displaystyle#1$\fi\ignorespaces}
\newcommand{\pushleft}[1]{\ifmeasuring@#1\else\omit$\displaystyle#1$\hfill\fi\ignorespaces}
\makeatother

\icmltitlerunning{Graph-based Time Series Clustering for End-to-End Hierarchical Forecasting}

\begin{document}

\twocolumn[
\icmltitle{Graph-based Time Series Clustering for End-to-End Hierarchical Forecasting}



\icmlsetsymbol{equal}{*}

\begin{icmlauthorlist}
\icmlauthor{Andrea Cini}{usi}
\icmlauthor{Danilo Mandic}{icl}
\icmlauthor{Cesare Alippi}{usi,polimi}
\end{icmlauthorlist}

\icmlaffiliation{usi}{The Swiss AI Lab IDSIA, Universit\`a della Svizzera italiana, Switzerland}
\icmlaffiliation{polimi}{Politecnico di Milano, Italy}
\icmlaffiliation{icl}{Imperial College London, United Kingdom}

\icmlcorrespondingauthor{Andrea Cini}{andrea.cini@usi.ch}
%\icmlcorrespondingauthor{Firstname2 Lastname2}{first2.last2@www.uk}

\icmlkeywords{time series forecasting, time series clustering, hierarchical forecasting, spatiotemporal data, graph neural networks}

\vskip 0.3in
]



\printAffiliationsAndNotice{}  % leave blank if no need to mention equal contribution

\begin{abstract}
Relationships among time series can be exploited as inductive biases in learning effective forecasting models. In hierarchical time series, relationships among subsets of sequences induce hard constraints~(hierarchical inductive biases) on the predicted values. In this paper, we propose a graph-based methodology to unify relational and hierarchical inductive biases in the context of deep learning for time series forecasting. In particular, we model both types of relationships as dependencies in a pyramidal graph structure, with each pyramidal layer corresponding to a level of the hierarchy. By exploiting modern -- trainable -- graph pooling operators we show that the hierarchical structure, if not available as a prior, can be learned directly from data, thus obtaining cluster assignments aligned with the forecasting objective. A differentiable reconciliation stage is incorporated into the processing architecture, allowing hierarchical constraints to act both as an architectural bias as well as a regularization element for predictions. Simulation results on representative datasets show that the proposed method compares favorably against the state of the art.
\end{abstract}
\section{Introduction}\label{sec:intro}

In most applications, collections of related time series can be organized and aggregated within a hierarchical structure~\cite{hyndman2011optimal}. One practical example is forecasting energy consumption profiles which can be aggregated at the level of individual households as well as at city, regional, and national scales~\cite{taieb2021hierarchical}. Similar arguments can be made for forecasting photovoltaic production~\cite{yang2017reconciling}, financial time series~\cite{athanasopoulos2020hierarchical}, and the influx of tourists~\cite{athanasopoulos2009hierarchical}, to name a few relevant application domains. By exploiting aggregation constraints, forecasts at different levels can be combined to obtain predictions at different resolutions. Similarly, coherency constraints can be used to regularize forecasts obtained for the different levels by considering \gls{fr} methods~\cite{hyndman2011optimal, wickramasuriya2019optimal, panagiotelis2023probabilistic}. Said differently, constraining forecasts at different levels to ``add up'' \textit{can} positively impact forecasting accuracy. Based on similar ideas, cluster-based aggregate forecasting methods learn to predict aggregates of clustered time series as an intermediate step for obtaining forecasts for the total aggregate~\cite{alzate2013improved, fahiman2017improving, cini2020cluster}. The idea underlying both approaches is that combining multiple forecasts reduces variance, an observation dating back to~\citet{bates1969combination}. In particular, \gls{fr} is a special case of forecast combinations~\cite{hollyman2021understanding}. 

Besides hierarchical constraints, correlated time series forecasting models can leverage relational inductive biases to predict any subset of the time series while sharing learnable parameters~\cite{cini2023taming}. Indeed, the combination of graph deep learning methods~\cite{bacciu2020gentle, stankovic2020data, bronstein2021geometric} and deep learning for time series~\cite{benidis2022deep} has led to state-of-the-art forecasting accuracy in several domains~\cite{jin2023survey, cini2023graphdeep}. Current state-of-the-art methods, however, are limited to processing the input data at a single spatial resolution which might cause propagation bottlenecks~\cite{alon2020bottleneck} and over-smoothing~\cite{rusch2023survey}. Graph pooling operators~\cite{grattarola2022understanding} enable \gls{gnn} architectures to learn how to cluster nodes and obtain hierarchical, higher-order, graph representations tailored to the task at hand~\cite{bianchi2023expressive}. Yet the application of learnable graph pooling operators and the combination of hierarchical and relational constraints are underexplored in graph-based forecasting.

To fill this void, this paper proposes a novel and comprehensive graph-based framework for hierarchical time series clustering and forecasting. Our approach unifies hierarchical time series processing, graph pooling operators, and graph-based neural forecasting methods. This results in a learning architecture for multi-step ahead forecasting operating at different levels of spatial resolution. Hierarchical and relational structures are embedded as inductive biases into the processing by exploiting neural message passing~\cite{gilmer2017neural} and graph pooling~\cite{grattarola2022understanding} operators. The proposed methodology, named \gls{method} can propagate representations along the hierarchical structure and ensure the coherency of predictions w.r.t.\ aggregation constraints. In particular, we focus on settings where the hierarchical structure is not given but learned directly from data. In this scenario, the forecast recombination mechanism is trained in a self-supervised manner, by exploiting the forecasting accuracy at different levels as a learning signal and the graph topology as a regularization mechanism. In other words, time series clusters are learned while maximizing the forecasting accuracy w.r.t.\ the corresponding aggregated time series. This provides an additional learning signal to the clustering procedure.

Our main novel contributions are:
\begin{itemize}%[leftmargin=1.5em]
    \item the introduction of a methodology to embed hierarchical constraints as inductive biases in graph-based forecasting architectures~(Sec.~\ref{sec:hierarchical-forecasting});
    \item a methodological framework, based on graph pooling, to learn a proper hierarchical structure directly from data by clustering the input time series~(Sec.~\ref{sec:end-to-end-cbaf});
    \item an end-to-end learning architecture incorporating the above components in a time series forecasting model~(Sec.~\ref{sec:hierarchical-forecasting}, \ref{sec:end-to-end-cbaf}, \ref{sec:forecasting-reconciliation}).
\end{itemize}
\gls{method} is extensively validated on relevant benchmarks~(Sec.~\ref{sec:experiments}). Besides achieving state-of-the-art forecasting accuracy, we show that our approach can be used as a self-supervised architecture to learn meaningful cluster assignments.

\section{Preliminaries}

This section introduces preliminary concepts and provides the problem settings. 

\paragraph{Graph-based spatiotemporal forecasting} Consider a set of $N$ univariate time series; $\vx_t^i \in \sR$ indicates the value observed at time step $t$ w.r.t.\ the $i$-th time series. The observation vector encompassing all the time series is analogously denoted by $\mX_t \in \sR^{N \times 1}$. Sequences of observations are indicated, e.g., as $\mX_{t:t+T}$ where the index $t:t+T$ refers to the time interval $[t, t+T)$. Available covariates can be encoded into a matrix $\mU_t\in\sR^{N\times d_u}$. 
We assume the considered time series to be spatially correlated; i.e., time series are not independent, but are instead characterized by functional dependencies affecting the temporal evolution of the system. Pairwise relationships among time series are encoded within a weighted adjacency matrix $\mA \in \sR^{N \times N}$ which is constant over time; the resulting attributed graph at time $t$ is denoted by the triple $\gG_t = \langle\mX_t, \mU_t, \mA\rangle$. Homogeneous sensors and static topology are assumed to ease the formalization of the problem; extensions beyond these settings are relatively straightforward, but outside of the paper's scope. 
The multi-step time series forecasting problem can then be modeled as the problem of predicting the $H$-step-ahead observations $\mX_{t:t+H}$ given a window of past data $\gG_{t-W:t}$ by minimizing some estimate of the forecasting error. We focus on \textit{point forecasts}, i.e., we do not model the uncertainty of the predictions. 

\paragraph{\Glspl{stgnn}} \Glspl{stgnn} are effective global time series forecasting models for the problem above. As a reference, we consider \gls{tts} architectures~\cite{ gao2021equivalence, cini2023graphdeep} with local learnable node embeddings~\cite{cini2023taming} where the input time series are processed by a temporal encoder followed by a stack of message-passing layers~\cite{gilmer2017neural} accounting for ``spatial'' dependencies such that
\begin{align}
    \vh^{i,0}_t &= \textsc{SeqEnc}\left(\vx^i_{t-W:t}, \vu^i_{t-W:t}, \vv^i\right),\\
    \vh^{i,l+1}_{t} &= {\update}^l\Big(\vh^{i,l}_{t}, \aggr_{j \in \mathcal{N}(i)}\Big\{{\msg}^l\big(\vh^{i,l}_{t}, \vh^{j,l}_{ t},a_{ji}\big)\Big\}\Big),\label{eq:mp}
\end{align}
where $\vv^i \subset \mV\in\sR^{N\times\ d_{e}}$ are the learnable node embeddings associated with the $i$-th node, ${\update}^l({}\cdot{})$ and ${\msg({}\cdot{})}^l$ indicate update and message functions at the $l$-th layer, respectively, which can be implemented by, e.g., \glspl{mlp}. $\textsc{SeqEnc}({}\cdot{})$ denotes a network encoding each input sequence along the temporal dimension, e.g., a \gls{rnn}, $\textsc{Aggr}\{{}\cdot{}\}$ is a permutation invariant aggregation function and $\mathcal{N}(i)$ refers to the set of neighbors of the $i$-th node, each associated to an edge with weight $a_{ji}$. In the following, the shorthand $\mH_t^{l+1} = \gnn_l(\mH_t^l,\mA)$ indicates a message-passing step w.r.t.\ the full node set and adjacency matrix $\mA$. Predictions can then be obtained by using any decoder, e.g., an \gls{mlp} followed by a linear readout for each prediction step.

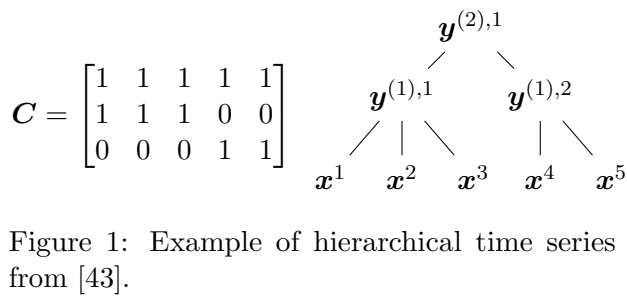
\begin{figure}%{l}{0.53\textwidth}
    %\centering
    \vspace{-.1cm}
\begin{minipage}{.23\textwidth}
\begin{equation*}
    \mC = \begin{bmatrix}
    1 &1&1&1&1\\
    1&1&1&0&0\\
    0&0&0&1&1%\\
    % 1 &0&0&0&0\\
    % 0&1&0&0&0\\
    % 0&0&1&0&0\\
    % 0&0&0&1&0\\
    % 0&0&0&0&1
    \end{bmatrix}
\end{equation*}
\end{minipage}
\begin{minipage}{.23\textwidth}
\begin{tikzpicture}[node distance=1.1cm]
  \node (top) at (0,0) {$\vy^{(2),1}$};
  \node (mid1) [below left of=top, node distance=1.3cm] {$\vy^{(1),1}$};
  \node (mid2) [below right of=top, node distance=1.3cm] {$\vy^{(1),2}$};
  \node (bot2) [below of=mid1] {$\vx^2$};
  \node (bot1) [left of=bot2, node distance=.95cm] {$\vx^1$};
  \node (bot3) [right of=bot2, node distance=.95cm] {$\vx^3$};
  \node (bot4) [below of=mid2] {$\vx^4$};
  \node (bot5) [right of=bot4, node distance=.95cm] {$\vx^5$};
  
  \draw (top) -- (mid1);
  \draw (top) -- (mid2);
  \draw (mid1) -- (bot2);
  \draw (mid1) -- (bot1);
  \draw (mid1) -- (bot3);
  \draw (mid2) -- (bot4);
  \draw (mid2) -- (bot5);
\end{tikzpicture}
\end{minipage}
    \caption{Example of hierarchical time series from~\cite{hyndman2018forecasting}.}
    \label{fig:hierachical-time-series}
\vspace{-.3cm}
\end{figure}

\paragraph{Hierarchical time series} In the hierarchical setting, the set of raw time series is augmented by considering additional sequences obtained by progressively aggregating those at the level below, thus building a pyramidal structure. 
In particular, \textit{bottom} observations~(raw time series) are denoted as $\mY^{(0)}_t = \mX_t$, while $\mY^{(k)}_t \in \sR^{N_k\times 1}$, with $k>0$, indicates values of $N_k$ series obtained by aggregating~(e.g., summing up) a partition of $\mY^{(k-1)}_t$. The full collection of both raw and aggregated observations is denoted by matrix $\mY_t \in \sR^{M\times 1}$, with $M = \sum_{k=0}^K N_k$, obtained by stacking the $\mY^{(k)}_t$ matrices vertically in decreasing order w.r.t.\ index $k$. In general, the level of the hierarchy is denoted as a superscript between parentheses.
The aggregation constraints can be encoded in an aggregation matrix $\mC \in \{0, 1\}^{(M-N)\times N}$ such that the $i$-th aggregate time series can be obtained as $\vy_t^i = \sum_{j=1}^N c_{ij}\vx^j_t$, i.e., by summing the bottom-level observations given the hierarchical constraints\footnote{Note the index $i$ does not refer to the level of the hierarchy but to the $i$-th element of the entire flattened collection $\mY_t$.}. Given the above, the following relationships hold:
\begin{align}
    \mY_t &= \begin{bmatrix} \mC\\ \mI \end{bmatrix} \mX_t, & \mQ \mY_t =\begin{bmatrix}\,\mI\,|\,-\mC\,\end{bmatrix}\mY_t &= \boldsymbol{0},\label{eq:coherency}
\end{align}
where $\mI$ indicates an identity matrix of appropriate dimensions and $|$ the concatenation operator.
Fig.~\ref{fig:hierachical-time-series} provides an example of a time series hierarchy with the associated aggregation matrix. A forecast $\widehat \mY_t$ is said to be \textit{coherent} if the equality constraints in Eq.~\ref{eq:coherency} holds, i.e., if $\mQ\widehat\mY_t =\boldsymbol{0}$. As discussed in the following, learning to forecast time series at different resolutions can act as an effective regularization mechanism, even when the hierarchical structure is not predefined.

\section{Graph-based Hierarchical Clustering and Forecasting}

\begin{figure*}[t]
    \centering
    \includegraphics[width=\textwidth]{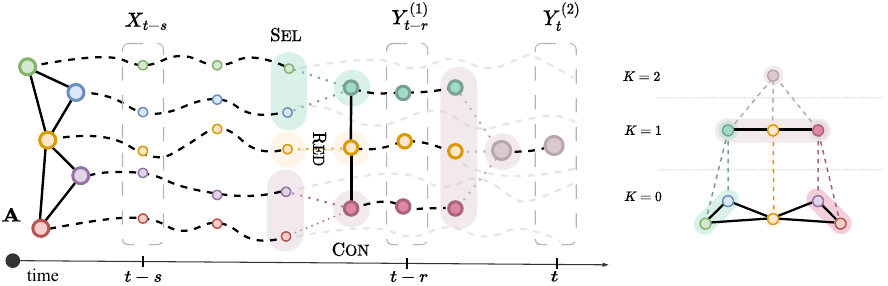}
    \caption{Time series with a hierarchical 
 relational structure. \textbf{(Left)} Graphical representation of hierarchical time series with graph-side information; SRC operators allow for modeling relationships among the time series in the hierarchy. \textbf{(Right)}~Pyramidal graph encompassing both hierarchical and relational dependencies; each pair of levels constitutes a bipartite graph.}\label{fig:architecture}
\end{figure*}

This section presents our approach to graph-based hierarchical time series forecasting. We start by discussing how to incorporate the hierarchical structure of the problem into a graph-based neural architecture~(Sec.~\ref{sec:hierarchical-forecasting}); then, we focus on our target setting and show how the hierarchical structure can be directly learned from data by exploiting trainable graph pooling operators~(Sec.~\ref{sec:end-to-end-cbaf}). Finally, we introduce an appropriate forecasting reconciliation mechanism to obtain forecasts coherent w.r.t.\ the learned hierarchy~(Sec.~\ref{sec:forecasting-reconciliation}). 

\subsection{Graph-based Hierarchical Forecasting}\label{sec:hierarchical-forecasting}

Embedding the hierarchical structure into the processing requires defining proper operators. In particular, we aim at designing a \textit{pyramidal} processing architecture where each layer corresponds to a level of the time series hierarchy and has its own topology, related to those at the adjacent layers by the hierarchical structure. To obtain such processing, operators have to be specified to control how information is propagated among and within the levels of the hierarchy; we exploit the connection to graph pooling for defining such operators within the \gls{src} framework~\cite{grattarola2022understanding}. In particular, we use \gls{src} building blocks as a high-level formalization of the operators required to perform clustering, aggregation, and graph rewiring at each level of the hierarchy. The three operators are defined as follows, by indicating as ${\mH_t^{(k)}\in\mR^{N_k\times d_h}}$ a feature matrix corresponding to representations at the $k$-th level of the hierarchy.
\begin{description}%[leftmargin=1.5em]
\item[Select] The selection operator $\select\emptyargs$ outputs a mapping from input nodes into supernodes~(i.e., clusters) given by the aggregation constraints at each level. The mapping can be encoded in a selection matrix ${\select(\mH^{(k)}_t, \dots) = \mS_k\in\{0, 1\}^{N_{k-1}\times N_{k}}}$ where $s_{ij}$ is equal to $1$ if and only if the $i$-th time series at level $k-1$ is mapped to the $j$-th aggregate at the $k$-th level. If the hierarchy is predefined, then the selection mechanism is given; conversely, learning a selection matrix is the key challenge for designing an end-to-end architecture and will be discussed in Sec.~\ref{sec:end-to-end-cbaf}.
\item[Reduce (and Lift)] The reduction function $\reduce\emptyargs$ aggregates node features and propagates information from the $k$-th level to the adjacent upper level in the hierarchy. Reduction can be obtained by summation, i.e., $\reduce(\mH^{(k-1)}_t, \mS^{(k)}) \doteq {\mS^{(k)}}^T\mH^{(k-1)}_t$, but other choices are possible. In practice, reduction is used in \gls{method} to propagate information along the pyramidal structure by aggregating node representations and implementing an inter-level message-passing mechanism~(see Eq.~\ref{eq:hierarchical-message-passing}). Similarly, we define the \textit{lift} operator as ${\lift(\mH^{(k+1)}_t, \mS^{(k)}) \doteq {\mS^{(k)}}\mH^{(k+1)}_t}$, i.e., as an upsampling the pooled graph to the original size obtained by mapping each supernode back to the aggregated nodes.
\item[Connect] The connect operator $\connect\emptyargs$ defines how the topology of the input graph is rewired after each aggregation step. There are several possible choices; we consider the rewiring where each pair of supernodes is connected by an edge with a weight obtained by summing weights of the edges from one subset to the other, i.e., ${\connect(\mS^{(k)}, \mA^{(k-1)}) \doteq {\mS^{(k)}}^T\mA^{(k-1)}\mS^{(k)}}$, where $\mA^{(k)}$ indicates the adjacency matrix w.r.t.\ $k$-th level.
\end{description}
These operators can be used to design neural processing architectures to match the inductive biases coming from the hierarchical structure. Fig.~\ref{fig:architecture} provides a graphical illustration of how these operators can be used to implement a hierarchical processing architecture. In particular, the figure shows subsequent applications of the selection, reduction and connection operators allow for operating on a progressively coarser graph structure accounting for higher-order dependencies. By exploiting the introduced operators, we can move from the reference architecture in Eq.~\ref{eq:mp} to a hierarchical \gls{tts} model operating as 
\begin{align}
     \vh^{(k),i,0}_t &= \textsc{SeqEnc}^{(k)}\left(\vy^{(k),i}_{t-W:t}, \vu^{(k),i}_{t-W:t}, \vv^{(k), i}\right),\label{eq:hier-seq-enc}\\
    {{\mZ}^{(k), l}_{t}} &= \gnn_{l}^{(k)}\left(\mH^{(k),l}_{t}, \mA^{(k)}\right),\label{eq:intra-mp}\\
    {{\mH}^{\scriptscriptstyle(k), l+1}_{t}} &= {\update}^{\scriptscriptstyle(k)}_l\Bigg(\mZ^{\scriptscriptstyle(k),l}_{t}, \underbrace{{\mS^{\scriptscriptstyle(k)}}^{\scriptscriptstyle T} \mZ^{\scriptscriptstyle(k-1),l}_{t}}_{\reduce^{\scriptscriptstyle(k)}}, \underbrace{\mS^{\scriptscriptstyle(k)} \mZ^{\scriptscriptstyle(k+1),l}_{t}}_{\lift^{\scriptscriptstyle(k)}}\Bigg).\label{eq:hierarchical-message-passing}
\end{align}
Eq.~\ref{eq:hier-seq-enc}, shows the temporal encoding step, Eq.~\ref{eq:intra-mp} refers to the intra-level propagation of messages, while Eq.~\ref{eq:hierarchical-message-passing} to the inter-level propagation; this needs further consideration to be fully appreciated. Matrix $\mH_t^{(k),l}$ indicates representations w.r.t.\ the $t$-th time step obtained at the $l$-th message-passing layer for time series at the $k$-th level of the hierarchy~(note the distinction between layers of message-passing and levels of the hierarchy). Compared to the model in Eq.~\ref{eq:mp}, the hierarchical constraints add further structure to the processing. As shown in Eq.\ref{eq:hier-seq-enc}, each time series is at first encoded along the temporal dimension by an encoder which can be either shared or different for each aggregation level. Then, representations are processed by a stack of layers propagating information within and among levels. As shown in Eq.~\ref{eq:hierarchical-message-passing}, the representations are updated at each step by an update function ${\update}_l^{(k)}\emptyargs$~(e.g., an \gls{mlp}) taking as an input (1) the output $\mZ_t^{(k),l}$ of a message-passing layer w.r.t.\ the graph topology at the $k$-th level~(Eq.~\ref{eq:intra-mp}), (2) aggregated features from the level $k-1$ and (3) the features corresponding to each node's supernode obtained by lifting $\mH^{(k+1),l}_{t}$. Learnable parameters may optionally be shared among the different levels of the hierarchy. Final predictions can be obtained by using an arbitrary readout, i.e., a standard \gls{mlp}, and by training the model to minimize the forecasting error w.r.t.\ all the time series as 
\begin{align}
    \hat \vy^{(k),i}_{t:t+H} = {\mlp}^{(k)}&\left(\vh^{(k),i,L}_t\right),\\  \mathcal{L}\left(\widehat \mY_{t:t+H}, \mY_{t:t+H}\right) &\doteq \Big\lVert\widehat \mY_{t:t+H} - \mY_{t:t+H}\Big\rVert_p^p,\label{eq:loss}
\end{align}
where $\mathcal{L}\emptyargs$ indicates the loss function and $p$ is equal to, e.g., $1$ or $2$. Note that the model is trained to make predictions for each level of the hierarchy. %at once which can act as a regularization method even if in the case where only base time series is the actual forecasting target. 
Representation at the different levels can capture patterns at different spatial scales, less apparent at fine-grained resolutions. Indeed, the aggregation and pooling operators increase the receptive field of each filter at each level of the hierarchy. Discussion on how to further regularize predictions given the hierarchical structure is postponed to  Sec.~\ref{sec:forecasting-reconciliation}.

\subsection{End-to-end Clustering and Forecasting}\label{sec:end-to-end-cbaf}

Learning a hierarchy and, consequently, a cluster-based forecasting architecture translates into learning a~(differentiable) parametrization of the selection operator. For this task, we provide a general probabilistic framework, based on modeling cluster assignments as realizations of a parametrized categorical distribution. Then, we briefly discuss the applicability of standard graph pooling methods from the literature at the end of the section. 

\paragraph{End-to-end clustering} Similarly to popular dense trainable graph pooling operators~\cite{bianchi2020spectral, ying2018hierarchical}, we parametrize the selection operator with a score matrix $\mPhi \in \sR^{N_{k-1} \times N_{k}}$, assigning a score $\phi_{ij}$ to each node-cluster pair. However, differently from previous works, we interpret such scores as~(unnormalized) log-probabilities, such that
\begin{align}
    \mPhi^{(k)} &= \mathcal{F}_{\boldsymbol{\psi}}\left(\mY^{(k-1)}_{t-W:t}, \mA^{(k-1)}, \mV^{(k-1)}\right),\notag\\
    \mS^{(k)} &\sim  P(\mS^{(k)}_{ij}=1) = \frac{e^{\phi^{(k)}_{ij}/\tau}}{\sum_j e^{\phi^{(k)}_{ij}/\tau}}\label{eq:learnable-pooling},
\end{align}
where $\tau$ is a temperature hyperparameter, while $\mathcal{F}_{\boldsymbol{\psi}}\emptyargs$ indicates a generic trainable function with trainable parameters $\boldsymbol{\psi}$. The conditioning on the input window $\mY_{t-W:t}^{(k-1)}$ can be dropped to obtain static cluster assignments; furthermore, depending on the dimensionality of the problem, the score matrix might also be parametrized directly as $\mPhi = \boldsymbol{\psi}$. Node embeddings and aggregates for the $k$-th level are then obtained through the reduction operator as ${\mV^{(k)} ={\mS^{(k)}}^T\mV^{(k-1)}}$ and ${\mY^{(k)}_t ={\mS^{(k)}}^T\mY_t^{(k-1)}}$, respectively. 
To differentiate through the sampling of $\mS^{(k)}$ we use the Gumbel softmax reparametrization trick~\cite{jang2017categorical, maddison2017concrete} followed by a discretization step to obtain hard cluster assignments via the straight-through gradient estimator~\cite{bengio2013estimating}. In practice, $\tau$ is set to $1$ at the beginning of training and is exponentially decayed towards $0$ at each training step. The above discretization step avoids soft cluster assignments that could lead to degenerate solutions given the loss in Eq.~\ref{eq:loss}. Uniform soft assignments are indeed likely to minimize the variance of the aggregate time series and thus the prediction error at levels $k > 0$. 

\paragraph{Graph-based regularization} To take the graph structure into account when learning the assignments, we exploit the min-cut regularization introduced by~\citet{bianchi2020spectral}, i.e., we add to the loss the term
\begin{align}
&\mathcal{L}^{c}\left(\mS_{\mu}^{\scriptstyle(k)},\mA^{(k-1)}\right) \doteq\label{eq:graph-reg}\\
&-\frac{\text{Tr}\left({{\mS_{\mu}^{\scriptstyle(k)}}^{\scriptscriptstyle T}\widetilde\mA^{\scriptstyle(k-1)}\mS_{\mu}^{(k)}}\right)}{\text{Tr}\left({\mS_{\mu}^{\scriptstyle(k)}}^{\scriptscriptstyle T}\widetilde\mD^{\scriptstyle(k-1)}\mS_{\mu}^{(k)}\right)} + \left\|\frac{{{\mS_{\mu}^{\scriptstyle(k)}}^{\scriptscriptstyle T}\mS_{\mu}^{\scriptstyle(k)}}}{\big\|{\mS_{\mu}^{\scriptstyle(k)}}^{\scriptscriptstyle T}\mS_{\mu}^{\scriptstyle(k)}\big\|_2} - \frac{\mI}{\sqrt{N_k}}\right\|_2\notag
\end{align}
where $\mS_{\mu}^{(k)} = \text{softmax}\big(\mPhi^{(k)}\big)$, $\widetilde\mD^{(k-1)}$ is the degree matrix of $\widetilde\mA^{(k-1)} \doteq \mD^{-\frac{1}{2}}\mA^{(k-1)}\mD^{-\frac{1}{2}}$, i.e., of the symmetrically normalized adjacency matrix. The first term in the equation is a continuous relaxation of the min-cut problem~\cite{dhillon2004kernel} incentivizing the formation of clusters that pool together connected components of the graph; the second term helps in preventing degenerate solutions by favoring orthogonal cluster assignments~\cite{bianchi2020spectral}. 

\paragraph{Training procedure} The training objective identified in Eq.~\ref{eq:loss} entails that the cluster assignments are learned to minimize the forecasting error w.r.t.\ both the bottom time series and aggregates. As a result, time series are clustered s.t.\ aggregates at all levels are easier to predict, thus providing a meaningful self-supervised learning signal. Intuitively, a signal will be easier to predict if characterized \emph{low intra-cluster variance}. At the same time, different levels in the hierarchy will benefit from reading information from diverse supernodes, thus favoring a \emph{high inter-cluster variance}. %Said differently, low-intra cluster variance leads to smooth aggregates which reduces the forecasting loss for $k\leq1$. At the same time, high inter-cluster variance is incentivized by the need to gather diverse information while learning~(as similarly happens with multi-head attention)/

\paragraph{Alternative pooling operators} Besides the clustering method described here, \gls{method} is compatible with any graph pooling approach from the literature~(see \citealt{grattarola2022understanding}). In particular, one might be interested in exploiting non-trainable graph pooling operators that obtain cluster assignments based on the graph topology only. The latter option becomes particularly attractive when obtaining predictions w.r.t.\ particular sub-graphs, or localized within specific connected components of the graph topology, is relevant for the downstream application. We discuss a selection of appealing methods from the literature in Sec.~\ref{sec:related-works} and refer to~\citet{grattarola2022understanding} for an in-depth discussion.

\subsection{Forecast Reconciliation}\label{sec:forecasting-reconciliation}

As mentioned in Sec.~\ref{sec:intro}, \gls{fr} allows for obtaining coherent forecast w.r.t.\ the hierarchical constraints~(Eq.~\ref{eq:coherency}). Furthermore, \gls{fr} can often have a positive impact on forecasting accuracy as reconciled forecasts are obtained as a combination of the predictions made at the different levels~\cite{hollyman2021understanding}. We follow \citet{rangapuram2021end} and embed a~(differentiable)~reconciliation step within the architecture as a projection onto the subspace of coherent forecasts. 

\paragraph{Forecast reconciliation} Given (trainable) selection matrices $\mS^{(1)}, \dots, \mS^{(K)}$ for each level of the hierarchy, the $\mQ$ matrix~(see Eq.~\ref{eq:coherency}) can be obtained as
\begin{align}
    &\mQ = \begin{bmatrix}\,\mI\,\Big|\,-\mC\,\end{bmatrix}=\\ 
    &=\begin{bmatrix}\,\mI\,\Big|\,-\begin{bmatrix}
        \begin{array}{c|c|c|c}
        \prod_{k=1}^{K} \mS^{\scriptstyle(k)} &  \prod_{k=1}^{K-1} \mS^{(\scriptstyle k-i)} & {\scriptstyle\cdots} & \mS^{\scriptstyle(1)}
    \end{array}%\end{bmatrix}^T. & \mQ &= \begin{bmatrix}
    \end{bmatrix}^{\scriptstyle T}\,    
    \end{bmatrix}.\notag
\end{align}
Then, raw predictions $\widehat \mY_t$ can be mapped into reconciled~(coherent) forecasts $\overline \mY_t$ through a projection onto the space of coherent forecasts~(i.e., the null space of $\mQ$). The projection matrix can be computed as
\begin{align}
    \mP &\doteq \mI - \mQ^T\left(\mQ\mQ^T\right)^{-1}\mQ, & \overline\mY_t = \mP\widehat\mY_t,
\end{align}
where $\mP$ is obtained by solving the constrained optimization problem $\min_{\mZ} \lVert \mZ - \widehat\mY_t\rVert_2$ s.t.\ ${\mQ\mZ =\boldsymbol{0}}$. Model parameters are then learned by minimizing the loss ${\mathcal{L}_f \doteq \mathcal{L}(\widehat\mY, \mY) + \mathcal{L}(\overline\mY, \mY)+ \lambda \mathcal{L}(\overline\mY, \widehat\mY)}$ where we omitted the time indices. Note that minimizing the regularization term $\mathcal{L}(\overline\mY, \widehat\mY)$ is equivalent to minimizing the distance between $\widehat \mY_{t:t+H}$ and the space of coherent forecasts. 
Unfortunately, computing the inverse of $\mQ\mQ^T$ incurs the cost $\bigO(M^3)$ in space and $\bigO(M^2)$ in time, which can be prohibitive for large time series collections. 
However, the solution is still practical for up to a few thousand nodes~(most practical applications), and the regularization term, computed as ${\mathcal{L}^{reg}(\widehat\mY, \lambda) \doteq \lambda \lVert\mQ\widehat\mY\rVert_2}$, can be used in the other cases as the only regularization. The above \gls{fr} method can be seamlessly integrated into our end-to-end forecasting framework, however, many possible alternatives could be considered here. The design of ad-hoc reconciliation methods for graph-based predictors is a promising research direction for future works~(see Sec.~\ref{sec:conclusion}).

\section{Related Work}\label{sec:related-works}

\paragraph{Hierarchical forecasting} Hierarchical forecasting is a widely studied problem in time series analysis~\cite{hyndman2018forecasting, hyndman2011optimal}. The standard approach consists of obtaining~(possibly independent) forecasts for a subset of time series in the hierarchy in the first stage and then, in a separate step, reconciling and combining them to obtain~(possibly coherent) predictions for the full hierarchy~\cite{hyndman2011optimal, taieb2019regularized, wickramasuriya2019optimal}. In particular, MinT~\cite{wickramasuriya2019optimal} allows for obtaining optimal reconciled forecasts given a set of unbiased $H$-step-ahead predictions and the covariance matrix of the associated residuals. Analogous reconciliation methods have also been developed for probabilistic forecasts~\cite{wickramasuriya2023probabilistic, taieb2017coherent, corani2021probabilistic}. End-to-end methods have been instead proposed in the context of deep learning for time series forecasting~\cite{benidis2022deep} by exploiting the hierarchical structure either as a hard~\cite{rangapuram2021end, zhou2023sloth, das2023dirichlet} or soft constraint~\cite{paria2021hierarchically, han2021simultaneously}. Notably, \citet{rangapuram2021end} incorporate the reconciliation step within the neural architecture as a differentiable convex optimization layer~\cite{agrawal2019differentiable} and obtain probabilistic forecasts by Monte Carlo sampling. None of these methods consider relational dependencies among and within the levels of the hierarchical structure.

\paragraph{Graph-based forecasting and graph pooling} Graph learning deep models have become popular in time series processing~\cite{li2018diffusion, cini2022filling, jin2023survey}. Graph pooling operators have been widely studied in \gls{gnn} models for i.i.d.\ data~\cite{grattarola2022understanding, bianchi2023expressive}, but their application to time series data is underexplored. Dense trainable pooling methods~\cite{ying2018hierarchical, bianchi2020spectral, hansen2022clustering} learn soft cluster assignment regularized by taking into account the graph structure. Sparse approaches, instead, produce hard cluster assignments usually learned by exploiting both the graph structure and a learned ranking on the nodes~\cite{bacciu2023pooling, gao2019graph}. Finally, non-trainable methods exploit a clustering of the nodes performed independently from the trained model~\cite{bianchi2020hierarchical, dhillon2007weighted}. Pyramidal graph-based architectures have been exploited in reservoir computing~\cite{bianchi2022pyramidal}. Graph neural networks has also been used to process temporal hierarchies by ~\citet{rangapuram2023coherent}. With regards to \glspl{stgnn}, hierarchical representations have been exploited in specific domains such as traffic analytics~\cite{yu2019st, guo2021hierarchical, hermes2022graph}, air quality monitoring~\cite{chen2021group}, financial time series~\cite{arya2023hierarchical}, and pandemic forecasting~\cite{ma2022hierarchical}. In particular, \citet{yu2019st} propose a spatiotemporal graph U-network~\cite{gao2019graph} where representations are pooled and then un-pooled to obtain a hierarchical processing of the time series.  However, most of the above methods rely on fixed cluster assignments; furthermore, none of them directly address the hierarchical time series forecasting problem by optimizing predictions at each level of the hierarchy to learn cluster assigments and taking into account coherency constraints.

\section{Experiments}\label{sec:experiments}

\gls{method} is validated over several settings considering forecasting benchmarks with no predefined hierarchical structure. In particular, we focus on validating of the proposed end-to-end clustering and forecasting architecture against relevant baselines and state-of-the-art architectures. We then provide a qualitative analysis of the learned time series clusters on datasets coming from sensor networks. Full details on the experimental setup are provided in Appendices~\ref{a:datasets} and \ref{a:baselines}, while Appendix~\ref{a:additiona-results} contains additional empirical results and sensitivity analyses.

\subsection{End-to-end Hierarchical Clustering and Forecasting}\label{sec:experiment-forecasting}

% Note use of \abovespace and \belowspace to get reasonable spacing
% above and below tabular lines.
\begin{table*}[t]
\caption{Forecasting performance on benchmark datasets (5 runs). Best result in \textbf{bold}, second best \underline{underlined}.}
\label{t:benchmarks}
\small
\setlength{\tabcolsep}{4pt}
\setlength{\aboverulesep}{0pt}
\setlength{\belowrulesep}{0pt}
\renewcommand{\arraystretch}{1.1}
\begin{center}
%\resizebox{\linewidth}{!}{%
\begin{tabular}{c | c | c c | c c  | c c | c c }
\toprule
 \multicolumn{2}{c|}{\multirow{2}{*}{\sc Models}} & \multicolumn{2}{c}{Metr-LA} & \multicolumn{2}{c}{PeMS-Bay} & \multicolumn{2}{c}{CER} & \multicolumn{2}{c}{AQI} \\
 \cmidrule{3-10}
 \multicolumn{2}{c|}{} & MAE & MRE (\%) & MAE & MRE (\%) & MAE & MRE (\%) & MAE & MRE (\%)\\
\midrule
\multicolumn{2}{c|}{RNN} & 3.543 {\tiny $\pm$ .005} & 6.134 {\tiny $\pm$ .008} & 1.773 {\tiny $\pm$ .001} & 2.839 {\tiny $\pm$ .001} & 4.57 {\tiny $\pm$ .00} & 21.65 {\tiny $\pm$ .01} & 14.00 {\tiny $\pm$ .03} & 21.84 {\tiny $\pm$ .05}\\
\multicolumn{2}{c|}{FC-RNN} & 3.566 {\tiny $\pm$ .018} & 6.174 {\tiny $\pm$ .031} & 2.305 {\tiny $\pm$ .006} & 3.690 {\tiny $\pm$ .009} & 7.13 {\tiny $\pm$ .02} & 33.77 {\tiny $\pm$ .11} & 18.33 {\tiny $\pm$ .11} & 28.59 {\tiny $\pm$ .18} \\
\midrule
\multicolumn{2}{c|}{GConv-TTS} & 3.071 {\tiny $\pm$ .008} & 5.317 {\tiny $\pm$ .015} & 1.584 {\tiny $\pm$ .006} & 2.536 {\tiny $\pm$ .009} & 4.12 {\tiny $\pm$ .02} & 19.50 {\tiny $\pm$ .08} & 12.30 {\tiny $\pm$ .02} & 19.20 {\tiny $\pm$ .03} \\
\multicolumn{2}{c|}{Diff-TTS} & 3.012 {\tiny $\pm$ .005} & 5.214 {\tiny $\pm$ .008} & 1.569 {\tiny $\pm$ .004} & 2.512 {\tiny $\pm$ .006} & 4.11 {\tiny $\pm$ .02} & 19.47 {\tiny $\pm$ .11} & 12.24 {\tiny $\pm$ .04} & 19.10 {\tiny $\pm$ .05} \\
\multicolumn{2}{c|}{Gated-TTS} & 3.027 {\tiny $\pm$ .008} & 5.240 {\tiny $\pm$ .013} & 1.582 {\tiny $\pm$ .006} & 2.533 {\tiny $\pm$ .009} & 4.13 {\tiny $\pm$ .01} & 19.54 {\tiny $\pm$ .06} & \underline{12.07 {\tiny $\pm$ .02}} & \underline{18.83 {\tiny $\pm$ .03}} \\
\multicolumn{2}{c|}{GUNet-TTS} & 3.057 {\tiny $\pm$ .016} & 5.292 {\tiny $\pm$ .028} & 1.575 {\tiny $\pm$ .006} & 2.522 {\tiny $\pm$ .010} & \underline{4.08 {\tiny $\pm$ .02}} & \underline{19.32 {\tiny $\pm$ .10}} & 12.25 {\tiny $\pm$ .03} & 19.11 {\tiny $\pm$ .05} \\
\midrule
\midrule
\multicolumn{2}{c|}{\gls{method}-TTS (C)} & 3.034 {\tiny $\pm$ .008} & 5.253 {\tiny $\pm$ .013} & \underline{1.567 {\tiny $\pm$ 0.005}} & \underline{2.508 {\tiny $\pm$ 0.008}} & 4.11 {\tiny $\pm$ .07} & 19.45 {\tiny $\pm$ .34} & 12.13 {\tiny $\pm$ .02} & 18.92 {\tiny $\pm$ .04}  \\
\multicolumn{2}{c|}{\gls{method}-TTS (D)} & \underline{3.009 {\tiny $\pm$ .005}} & \underline{5.209 {\tiny $\pm$ .008}} & \textbf{1.566 {\tiny $\pm$ .005}} & \textbf{2.506 {\tiny $\pm$ .008}} & 4.12 {\tiny $\pm$ .06} & 19.49 {\tiny $\pm$ .30} & 12.10 {\tiny $\pm$ .01} & 18.88 {\tiny $\pm$ .02}  \\
\multicolumn{2}{c|}{\gls{method}-TTS (G)} & \textbf{3.007 {\tiny $\pm$ .009}} & \textbf{5.205 {\tiny $\pm$ .016}} & 1.568 {\tiny $\pm$ .008} & 2.510 {\tiny $\pm$ .013} & \textbf{4.05 {\tiny $\pm$ .01}} & \textbf{19.20 {\tiny $\pm$ .03}} & \textbf{12.02 {\tiny $\pm$ .04}} & \textbf{18.75 {\tiny $\pm$ .06}} \\
\bottomrule
\end{tabular}%
%}
\end{center}
% \vskip -.1in
\end{table*}

\begin{table}[t]
% \vspace{-0.15cm}
\caption{Results on traffic datasets (5 runs). Best results in \textbf{bold}, second best \underline{underlined}. %MAE@X indicates the MAE w.r.t.\ the X-minutes-ahead forecast.
}
% \vspace{-0.3cm}
\label{t:traffic}
\small
\setlength{\tabcolsep}{4pt}
\setlength{\aboverulesep}{0pt}
\setlength{\belowrulesep}{0pt}
\renewcommand{\arraystretch}{1.1}
\begin{center}
% \resizebox{\linewidth}{!}{%
\begin{tabular}{c | c | c c c}
\toprule
 \multicolumn{2}{c|}{\multirow{2}{*}{\sc Models}} & \multicolumn{3}{c}{MAE}\\
 \cmidrule{3-5}
 \multicolumn{2}{c|}{} & 15 min. & 30 min. & 60 min. \\
\midrule
\multirow{7}{*}{\scalebox{0.94}{\rotatebox[origin=c]{90}{\gls{la}}}}&\multicolumn{1}{c|}{DCRNN} & 2.82 {\tiny $\pm$ .00} & 3.23 {\tiny $\pm$ .01} & 3.74 {\tiny $\pm$ .01} \\
&\multicolumn{1}{c|}{GWNet} & 2.72 {\tiny $\pm$ .01} & 3.10 {\tiny $\pm$ .02} & 3.54 {\tiny $\pm$ .03} \\
&\multicolumn{1}{c|}{Gated-GN} & 2.72 {\tiny $\pm$ .01} & \underline{3.05 {\tiny $\pm$ .01}} & \underline{3.44 {\tiny $\pm$ .01}}\\
&\multicolumn{1}{c|}{SGP} & \underline{2.69 {\tiny $\pm$ .00}} & \underline{3.05 {\tiny $\pm$ .00}} & 3.45 {\tiny $\pm$ .00} \\
\cmidrule{2-5}
&\multicolumn{1}{c|}{\gls{method} (T)} & \textbf{2.68 {\tiny $\pm$ .01}} & \textbf{3.02 {\tiny $\pm$ .01}} & \textbf{3.40 {\tiny $\pm$ .01}} \\
\cmidrule{2-5}
\cmidrule{2-5}
&\multicolumn{1}{c|}{No rel.\ prop.} & 2.80 {\tiny $\pm$ .01} & 3.14 {\tiny $\pm$ .01} & 3.47 {\tiny $\pm$ .02} \\
&\multicolumn{1}{c|}{No hier.\ prop.} & \textbf{2.68} {\tiny $\pm$ .01} & 3.03 {\tiny $\pm$ .02} & 3.43 {\tiny $\pm$ .02} \\
\midrule
\midrule
\multirow{7}{*}{\scalebox{0.94}{\rotatebox[origin=c]{90}{\gls{bay}}}}&\multicolumn{1}{c|}{DCRNN} &  1.36 {\tiny $\pm$ .00} & 1.71 {\tiny $\pm$ .00} & 2.08 {\tiny $\pm$ .01} \\
&\multicolumn{1}{c|}{GWNet} &  \underline{1.31 {\tiny $\pm$ .00}} & 1.64 {\tiny $\pm$ .01} & 1.94 {\tiny $\pm$ .01} \\
&\multicolumn{1}{c|}{Gated-GN} & 1.32 {\tiny $\pm$ .00} & 1.63 {\tiny $\pm$ .01} & 1.89 {\tiny $\pm$ .01} \\
&\multicolumn{1}{c|}{SGP} & \textbf{1.30 {\tiny $\pm$ .00}} & \textbf{1.60 {\tiny $\pm$ .00}} & \underline{1.88 {\tiny $\pm$ .00}} \\
\cmidrule{2-5}
&\multicolumn{1}{c|}{\gls{method} (T)} & \underline{1.31 {\tiny $\pm$ .00}} & \underline{1.61 {\tiny $\pm$ .00}} & \textbf{1.87 {\tiny $\pm$ .00}} \\
\cmidrule{2-5}
\cmidrule{2-5}
&\multicolumn{1}{c|}{No rel.\ prop.} & 1.32 {\tiny $\pm$ .00} & 1.63 {\tiny $\pm$ .00} & 1.88 {\tiny $\pm$ .01} \\
&\multicolumn{1}{c|}{No hier.\ prop.} & 1.31 {\tiny $\pm$ .00} & 1.63 {\tiny $\pm$ .00} & 1.89 {\tiny $\pm$ .00} \\
\bottomrule
\end{tabular}%
% }
\end{center}
\end{table} 

The empirical evaluation was set up by considering the following benchmarks and baselines.

\paragraph{Benchmarks} We consider the multistep-ahead forecasting task and benchmark data coming from medium-sized sensor networks~(hundreds of nodes). In particular, the benchmark consists of four datasets in total and includes two datasets from the traffic forecasting literature~(\textbf{\gls{la}} and \textbf{\gls{bay}}, \citealt{li2018diffusion}), one dataset of air quality measurements~(\textbf{\gls{air}}, \citealt{zheng2015forecasting}) and a collection of energy consumption profiles~(\textbf{\gls{cer}}, \citealt{cer2016cer}). Each dataset consists of correlated time series with graph-side information; no explicit prior hierarchical structure is given. We follow the setup of~\cite{cini2023taming}, by adopting the same splits for training, validation, and testing and the same procedure followed of previous works to extract a graph topology for each dataset~\cite{wu2019graph, cini2022filling}. Similarly, the length of the input window and forecasting horizon for each dataset are set according to related works~\cite{li2018diffusion, cini2023taming} as detailed in the Appendix~\ref{a:datasets}. We use the \textit{mean absolute error}~(MAE) and the \textit{mean relative error}~(MRE) as performance metrics. 

\paragraph{Baselines} To carry out meaningful comparisons we select a reference \gls{tts} architecture~\cite{cini2023taming, gao2021equivalence}~(see Eq.~\ref{eq:mp}) obtained by stacking an node-wise temporal encoder implemented by an \gls{rnn}, two \gls{gnn} layers, and an \gls{mlp} readout as
\begin{equation*}
    \text{\gls{rnn}}[d_h]-\text{MP}[d_h]-\text{MP}[d_h]-\text{FC}[d_h]-\textsc{Lin}[H]
\end{equation*}
where MP indicates a generic message-passing block, FC indicates a dense fully connected layer, and $\textsc{Lin}(H)$ is a linear layer with an output size corresponding to the forecasting horizon. The number of neurons in each layer is indicated as $d_h$. Learnable node embeddings~\cite{cini2023taming} are concatenated to the input before both the recurrent encoder and after the message-passing layers. We compare the performance of different message-passing schemes commonly used in state-of-the-art graph-based forecasting architectures. In particular, the considered alternatives include the standard graph convolution~(\textbf{GConv-TTS}, \citealt{kipf2017semi}), the bidirectional diffusion convolution operator~(\textbf{Diff-TTS}, \citealt{li2018diffusion}), a more advanced gated message-passing scheme~(\textbf{Gated-TTS},~\citealt{cini2023taming}), and a hierarchical Graph U-Net~(\textbf{GUNet-TTS}, \citealt{gao2019graph}). We use a standard \textbf{GRU}~\cite{cho2014properties} as sequence encoder for all the baselines. Finally, we denote by \textbf{FC-RNN} the baseline which considers the input sequences as a single multivariate time series and by $\textbf{\gls{rnn}}$ the global univariate model. The number of neurons $d_h$ is selected for each dataset on the validation set~(more details in the appendix), while the other hyperparameters are kept fixed among baselines~(see Appendix~\ref{a:baselines}). The \textbf{\gls{method}-TTS} model is implemented following the above template and Eq.~\ref{eq:hier-seq-enc}--\ref{eq:hierarchical-message-passing}. Notably, the only architectural difference w.r.t.\ the baselines is the addition of a hierarchical propagation step after each message-passing layer and readouts for each level of the hierarchy. \gls{method} is trained end-to-end as to minimize the forecasting error w.r.t.\ the aggregates corresponding to the learned clusters. For this experiment, we use a static learnable hierarchical structure with $3$ levels consisting of raw time series at the bottom, $20$ supernodes in the middle level, and the total aggregate as the single time series at the top level. Selection matrices are learned directly by parametrizing the associated log-probabilities with tables of trainable parameters.

\begin{figure*}[t]
    \centering
    \begin{subfigure}[b]{.93\textwidth}
        \includegraphics[width=\linewidth]{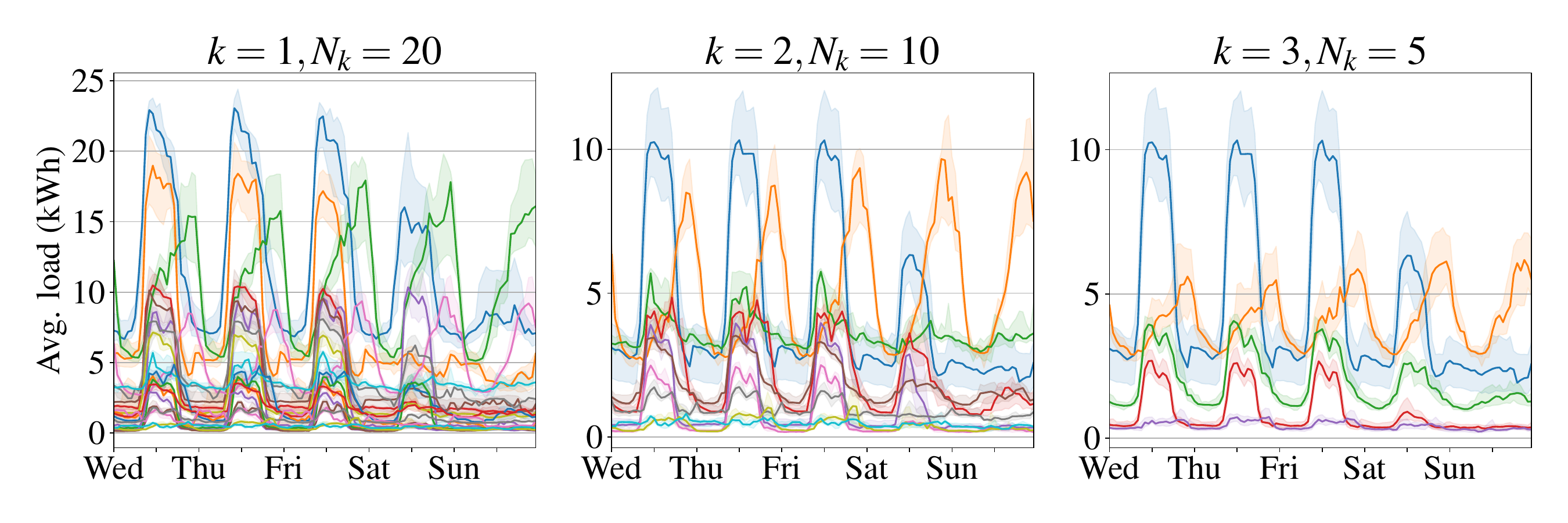}
        \caption{\gls{cer} dataset.}
        \label{fig:cer_clusters}    
    \end{subfigure}
    \begin{subfigure}[b]{0.93\textwidth}
        \includegraphics[width=\linewidth]{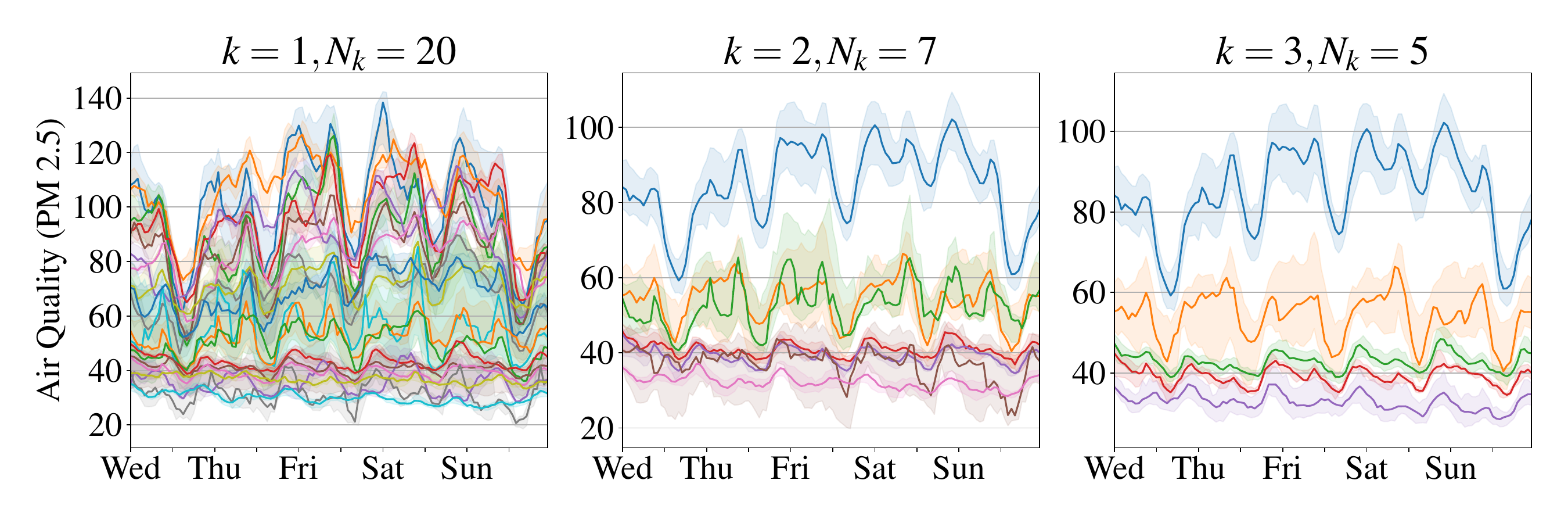}
        \caption{\gls{air} dataset.}
        \label{fig:air_clusters}
    \end{subfigure}
    \caption{Hierarchical cluster assignments learned by \gls{method} on $2$ benchmark datasets. The models have been trained with a $5$-level hierarchy and plots show, from left to right, the median for the clusters corresponding to levels from $1$ to $3$. The shaded areas correspond to $0.6$ and $0.4$ quantiles.}
    \label{fig:clusters}
\end{figure*}

\paragraph{Results} Tab.~\ref{t:benchmarks} show the results of the extensive empirical evaluation. We report \gls{method} forecasting accuracy w.r.t.\ $3$ different message-passing schemes; in particular, (C), (D), and (G) indicate respectively the standard graph convolution, the diffusion convolution operator and the gated message-passing operator cited above. \Gls{method} variants are among the best-performing methods in all the considered settings. Notably, hierarchical forecasting does not only act as self-supervision to learn cluster assignments but also provides a positive inductive bias that results -- on average -- in improved forecasting accuracy w.r.t.\ the flat architectures. Conversely, the GUNet baseline provides a comparison with a standard hierarchical message-passing architecture which, in this case, underperforms.

\paragraph{Comparison against the state of the art} Next, we perform an additional experiment by taking advantage of the popularity of \gls{la} and \gls{bay} as traffic forecasting benchmarks and compare \gls{method} against specialized state-of-the-art architectures. We consider the following baselines from the literature: 1) \textbf{DCRNN}~\cite{li2018diffusion}, i.e., a recurrent architecture; 2) Graph WaveNet~(\textbf{GWNet}, \citealt{wu2019graph}), i.e., a popular time-and-space graph convolutional model;  3) \textbf{Gated-GN}~\cite{satorras2022multivariate}, i.e., a gated message-passing architecture operating on a fully connected graph; 4) \textbf{SGP}~\cite{cini2023scalable}, i.e., a scalable architecture exploiting a randomized spatiotemporal encoder. In this context, we tuned the \gls{method} architecture by simply adding residual connections and using a deeper \gls{mlp} decoder; the tuned architecture is designeted as \gls{method}~(T). The simulation results for multistep-ahead forecasting in the traffic datasets, provided in Tab.~\ref{t:traffic}, show that \gls{method} can achieve state-of-the-art forecasting accuracy. Additionally, the same table reports an ablation study of the proposed architecture. In particular, we consider two variants of the model: the first is characterized by the removal of all the message-passing layers, while the second does not perform any propagation of the learned representations through the learned hierarchy. Results show that both aspects have a significant impact on forecasting accuracy.

\subsection{Cluster Analysis}\label{sec:experiment-end-to-end}

We analyze clusters extracted by \gls{method} on the \gls{cer} and \gls{air} datasets. 
Ideally, we would like to cluster customers w.r.t.\ their consumption patterns in the first case, and to partition air quality monitoring stations w.r.t.\ the different dynamics and regions of the dataset. As discussed in Sec.~\ref{sec:end-to-end-cbaf}, \gls{method} learns the cluster assignments by minimizing the forecasting error at each level of the hierarchy end-to-end. This form of self-supervision rewards, then, the formation of clusters that result in aggregates that are easy to predict and that, at the same time, are formed by taking the graph structure into account~(Eq.~\ref{eq:graph-reg}). We configure \gls{method} to learn $3$ hierarchical cluster assignments and show the result of the procedure in Fig.~\ref{fig:clusters}. In both scenarios, \gls{method} extracts meaningful clusters with aggregates exhibiting different patterns. In particular, each level corresponds to progressively smoother dynamics. Appendix~\ref{a:clusters} provides a spatial visualization of the clustered nodes for the AQI dataset.

\section{Conclusions}\label{sec:conclusion}

We introduced the \acrlong{method}, a methodological framework unifying relational and hierarchical inductive biases in deep learning architectures for time series forecasting. \gls{method} has been designed to learn hard cluster assignments end-to-end, by taking the graph structure into account and minimizing the forecasting error w.r.t.\ the resulting aggregates and bottom-level time series. Performance on relevant benchmarks supports the validity of the approach which, as we show, can also learn meaningful hierarchical cluster assignments. 

\paragraph{Future works} There are many possible extensions to the framework, which can be seen as a starting point for several specific studies and research directions. Future works might focus on the clustering aspect and investigate additional auxiliary objectives to provide more supervision to the procedure. Alternative reconciliation strategies should be assessed as well, together with their impact on the learned cluster assignments and forecasting accuracy. Future research could also apply \gls{method}-like methods to settings where the hierarchical constraints are predefined. The sensitivity of the approach to the number of input time series and observations might also be further explored; notably, the number of time series usually considered in graph-based forecasting is higher than those considered in standard hierarchical forecasting benchmarks. Finally, extensions of the framework to multivariate, heterogenous, and irregularly sampled time series would make the approach applicable to additional relevant and practical application domains. 

% \clearpage

\section*{Acknowledgements}

This work was partially supported by the Swiss National Science Foundation project FNS 204061: \emph{Higher-Order Relations and Dynamics in Graph Neural Networks}. The authors thank Filippo Maria Bianchi and Ivan Marisca for the useful feedback and discussions and Stefano Imoscopi for helping with data pre-processing.
\section*{Impact Statement}

This paper presents work whose goal is to advance the field of machine learning and time series forecasting. There are many potential societal consequences of our work, none of which we feel must be specifically highlighted here.

\bibliographystyle{plainnat}
\bibliography{bibliography}

\clearpage

\onecolumn
\appendix
\section*{Appendix}

This appendix provides additional details on the setup  and datasets used for the experiments presented in the paper, as well ass additional empirical results. 

\section{Hardware and software platforms}\label{a:exp}

Experimental setup and baselines  have been developed with Python~\cite{rossum2009python} by relying on the following open-source libraries:
\begin{itemize}
    \item numpy~\cite{harris2020array};
    \item PyTorch~\cite{paske2019pytorch};
    \item PyTorch Lightning~\cite{Falcon_PyTorch_Lightning_2019};
    \item PyTorch Geometric~\cite{fey2019fast};
    \item Torch Spatiotemporal~\cite{Cini_Torch_Spatiotemporal_2022}.
\end{itemize}

Experiments were run on a server equipped with AMD EPYC 7513 CPUs and NVIDIA RTX A5000 GPUs. The code for reproducing the computational experiments is available online\footnote{\url{https://github.com/andreacini/higp}}.

\section{Datasets}\label{a:datasets}

\begin{table*}[ht]
\caption{Statistics of datasets used in the experiments.}
\label{t:datasets}
\vskip 0.1in
\setlength{\aboverulesep}{0pt}
\setlength{\belowrulesep}{0pt}
\renewcommand{\arraystretch}{1.2}
\centering
\begin{small}
\begin{tabular}{l|c c c c}
\toprule
 \sc Datasets & Time steps & Nodes & Edges & Type\\
\toprule
\gls{la} & 34,272 & 207 & 1515 & Directed\\
\gls{bay} & 52,128 & 325 & 2369 & Directed\\
\gls{cer} & 25,728 & 485 & 4365 & Directed\\
\gls{air} & 8,760 & 437 & 2699 & Undirected\\
\bottomrule
\end{tabular}
\end{small}
\end{table*}

We use the same spatiotemporal forecasting benchmarks of~\cite{cini2023taming}, which consist of the following datasets.
\begin{description}[leftmargin=1.5em]
    \item[\textbf{\gls{la}}] The \gls{la} dataset~\cite{li2018diffusion} consists of measurements from loop detectors in the Los Angeles County Highway. 
    \item[\textbf{\gls{bay}}] The \gls{bay} dataset~\cite{li2018diffusion}, contains traffic speed measurements analogous to those of \gls{la} and acquired in the San Francisco Bay Area. 
    \item[\textbf{\gls{cer}}] The \gls{cer} dataset~\cite{cer2016cer} consists of a collection of load profiles~(i.e., energy consumption measurements) aggregated into $30$-minutes intervals, recorded by $485$ smart meters in Irish small and medium-sized enterprises. The dataset has been introduced as a benchmark for graph-based time series processing in~\cite{cini2022filling}.
    \item[\textbf{\gls{air}}] The \gls{air} dataset~\cite{zheng2015forecasting} collects hourly measurements of the PM2.5 pollutant from $437$ air quality monitoring stations spread over $43$ Chinese cities. Similarly to \gls{cer}, \gls{air} has been introduced as a benchmark for graph-based processing in \cite{cini2022filling}.
\end{description}
All of the above datasets are either openly available ~(\gls{la}, \gls{bay}, \gls{air}) or obtainable free of charge for research purposes~(\gls{cer}\footnote{\url{https://www.ucd.ie/issda/data/commissionforenergyregulationcer/}}). Tab.~\ref{t:datasets} provides relevant statistics on the considered datasets. For each dataset, we obtain the corresponding adjacency matrix and exogenous variables by following previous works~\cite{cini2022filling, li2018diffusion, cini2023taming}. Following~\citet{cini2023taming}, datasets are split into windows of $W$ time steps and the models are trained to predict the subsequent $H$ observations. Window size $W$ and forecasting horizon $H$ are respectively set as $W=12$ and $H=12$ for \gls{la} and \gls{bay}, $W=48, H=6$ for \gls{cer}, and $W=24, H=3$ for \gls{air}. Training, validation, and testing data are respectively obtained with a $70\%/10\%/20\%$ sequential split. Conversely, for \gls{air}, we use the same data splits of~\cite{yi2016st}.

\section{Baselines and hyperparameters}\label{a:baselines}

\subsection{Reference architectures} 

As discussed in Sec.~\ref{sec:experiments}, the main empirical results of the paper~(Tab.~\ref{t:benchmarks}), were obtained by considering, for all the baselines, a template TTS architecture which can be schematically described as follows:
\begin{align}
    \vh^{i,0}_t &= \text{GRU}\left(\vx_{t-W:t}^{i}, \vu_{t-W:t}^i, \vv^i\right),\\
    \mH^1_t &= \text{GNN}_1\left(\mH^0_t, \mA\right),\\
    \mH^2_t &= \text{GNN}_2\left(\mH^1_t, \mA\right),\\
    \hat{\vx}_{t+h}^i &=\mW_h \xi\left(\mW_{fc}\left[\vh_t^{i,2}|\vv_i\right] + \vb_{fc}\right) + \vb_h, \qquad h=0,1,\dots,H-1,
\end{align}
with $\xi\emptyargs$ being the ELU activation function~\cite{clevert2015fast}, $\mW_h \in \sR^{1 \times d_h}$, $\mW_h \in \sR^{d_h \times d_h}$, $\vb_{h} \in \sR$, $\vb_{fc} \in \sR^{d_h}$ denoting learnable weights, \textit{GRU} and \textit{GNN} indicating respectively a gated recurrent temporal encoder~\cite{cho2014properties} and a generic message-passing layer~(implemented differently for each baseline).
For \gls{method}, the template was modified to account for the hierarchical structure as discussed in Sec.~\ref{sec:hierarchical-forecasting}. Similarly, for the GUNet baselines the template was modified to take into account the pooling and lifting operations. For the tuned version of \gls{method} we simply added skip connections and used a deeper readout.

\subsection{Hyperparameters and training details} 

We trained each model with early stopping on the validation set and a batch size of $64$ samples for a maximum of $200$ epochs each of $300$ batches maximum. We used the Adam optimizer with an initial learning rate of $0.003$ reduced by a factor $\gamma=0.25$ every $50$ epochs. The number of neurons $d_h$ in the layers of each model was set to $64$ or $32$ based on the validation error on each dataset. For \gls{method}, the regularization coefficient $\lambda$ was tuned and set to $0.25$ based on the validation error on the \gls{la} dataset and simply rescaled for the other datasets to take into account the different magnitude of the input. As discussed in Sec.~\ref{sec:experiments}, we used a $3$-level hierarchy with $20$ super-nodes in the middle level and a single super-node~(the total aggregate) at the top level. Intra-level spatial propagation was performed only at the base level. For the Diff-TTS baseline, the order of the diffusion convolution was set to $k=2$, while the pooling factor for the GUNet was set to $p=0.1$. For what concerns the experimental results in Tab.~\ref{t:traffic}, for each baseline we used the hyperparameters of the original papers and the open-source implementation provided by \citet{cini2023scalable}. Hyperparameters for \gls{method}~(T) were obtained by tuning the model on the validation set of both datasets separately. 

\section{Additional results}\label{a:additiona-results}

\subsection{Sensitivity analyses}\label{a:sensitivity}

\paragraph{Hierarchy size} We ran a sensitivity analysis to assess the impact of the number of clusters and levels in the hierarchy. In particular, we ran the following experiment on the \gls{cer} dataset using a simplified model~(with $32$ hidden units in each layer) to test different configurations. Each configuration addresses a hierarchy with different numbers of levels and clusters. The results in Tab.~\ref{t:sensitivity-hierarchy} show how forecasting accuracy varies across configurations and that these hyperparameters should be tuned on the task at hand. However, as shown in Tab.~\ref{t:benchmarks}, we observed that using a simple hierarchy with $3$ levels and a small number of clusters is sufficient to outperform flat predictors consistently.

\begin{table}[h]
% \vspace{-0.15cm}
\caption{Sensitivity analysis comparing different clustering hyperparameters in terms on MAE on \gls{cer}~(4 runs). The number of time series in each level are indicated between parentheses for each setting. 
}
\vspace{-0.3cm}
\label{t:sensitivity-hierarchy}
\renewcommand{\arraystretch}{1.1}
\setlength{\tabcolsep}{4pt}
\setlength{\aboverulesep}{0pt}
\setlength{\belowrulesep}{0pt}
\begin{center}
% \resizebox{\linewidth}{!}{%
\begin{tabular}{ c | c c }
\toprule
 \multicolumn{1}{c|}{\multirow{2}{*}{Levels}} & \multicolumn{2}{c}{Hierarchy}\\
 \cmidrule{2-3}
 \multicolumn{1}{c|}{} & Sparse & Dense. \\
\midrule
\multirow{2}{*}{3 levels} & 4.24 {\tiny $\pm$ .02} & 4.26 {\tiny $\pm$ .02}  \\
 & (N, 20, 1) & (N, 100, 1)  \\
\midrule
\multirow{2}{*}{4 levels} & 4.22 {\tiny $\pm$ .02} & 4.23 {\tiny $\pm$ .02}  \\
 & (N, 20, 4, 1) & (N, 100, 50, 1)  \\
\midrule
\multirow{2}{*}{5 levels} & 4.24 {\tiny $\pm$ .01} & {4.23 {\tiny $\pm$ .02}} \\
 & (N, 100, 20, 4, 1) & (N, 200, 100, 50, 1)  \\
\bottomrule
\end{tabular}%
% }
\end{center}
\end{table}

\paragraph{Reconciliation strategy} The reconciliation procedure should be considered as a hyperparameter of the approach. In this regard, we ran a sensitivity analysis on the \gls{cer} dataset considering the \gls{method} model equipped with gated graph convolutions. Results are shown in Fig.~\ref{t:sensitivity-rec}: \textit{Reconciled}, \textit{Only loss} and \textit{Fit only base} indicate respectively the full model including the hard reconciliation step, the model simply minimizing accuracy at all levels at the same time, and  the model where the loss is computed only w.r.t.\ the base time series. Results further confirm that reconciliation and hierarchical biases can improve forecasting accuracy.

\begin{table}[h]
% \vspace{-0.15cm}
\caption{Sensitivity analysis comparing reconciliation strategies in terms on MAE on \gls{cer}~(4 runs).
}
\vspace{-0.3cm}
\label{t:sensitivity-rec}
\renewcommand{\arraystretch}{1.1}
\setlength{\tabcolsep}{4pt}
\setlength{\aboverulesep}{0pt}
\setlength{\belowrulesep}{0pt}
\begin{center}
% \resizebox{\linewidth}{!}{%
\begin{tabular}{ c | c c }
\toprule
 % \multicolumn{1}{c|}{\multirow{2}{*}{Method}} \\
 % \cmidrule{2-3}
 \multicolumn{1}{c|}{Method} & MAE & MRE \\
\midrule
\multirow{1}{*}{Reconciled} & 4.06 {\tiny $\pm$ .01} & 19.202 {\tiny $\pm$ .039}  \\
\midrule
\multirow{1}{*}{Only loss} & 4.08 {\tiny $\pm$ .01} & 19.309 {\tiny $\pm$ .034}  \\
\midrule
\multirow{1}{*}{Fit only base} & 4.07 {\tiny $\pm$ .01} & {19.245 {\tiny $\pm$ .052}} \\
\bottomrule
\end{tabular}%
% }
\end{center}
\end{table}

\subsection{Cluster analysis}\label{a:clusters} 

\begin{figure}[ht]
    \centering
    \includegraphics[width=0.7\textwidth]{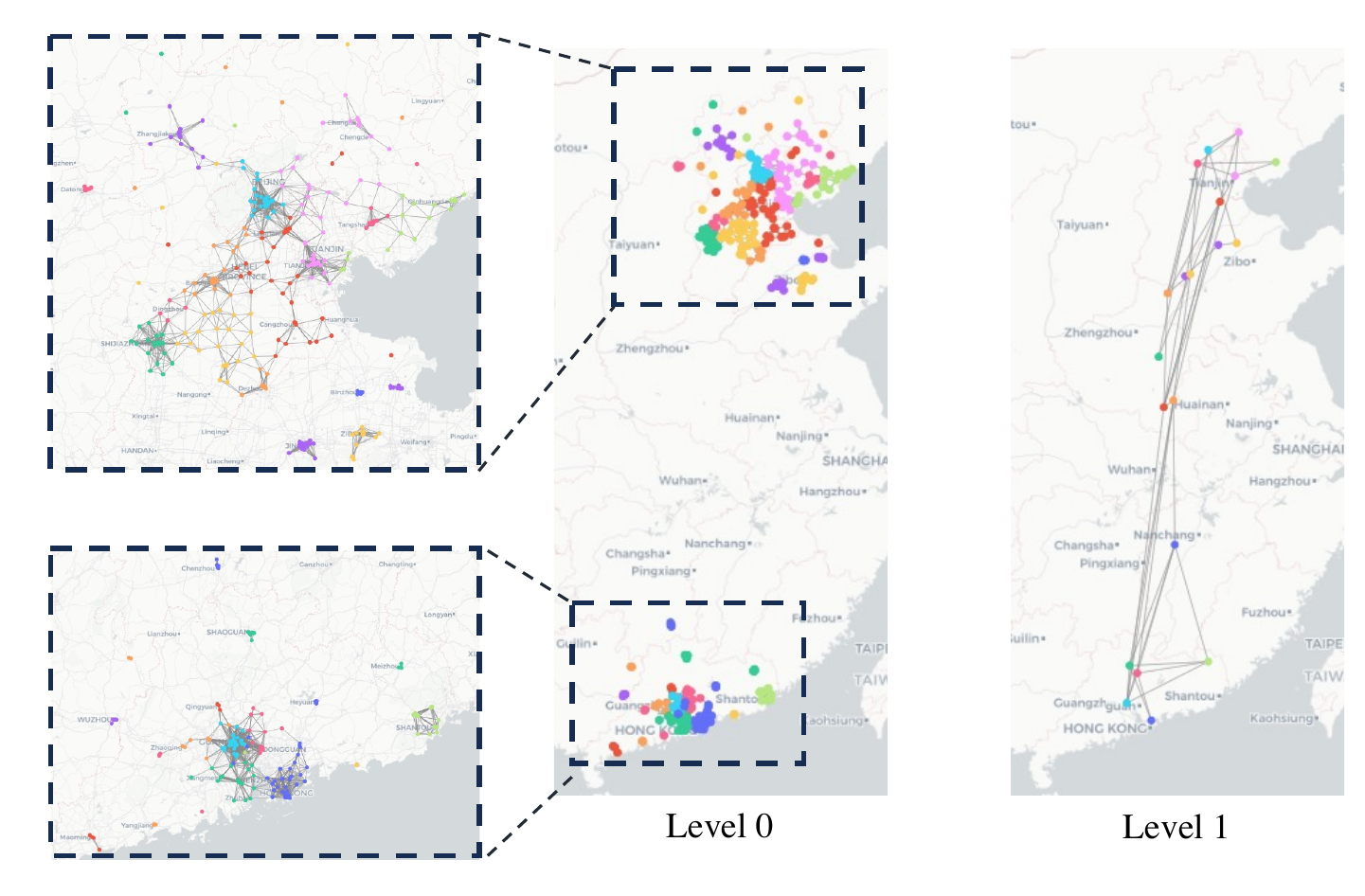}
    \vspace{-.3cm}
    \caption{Visualizations of clustered nodes in the AQI dataset.}
    \label{fig:aqi-spatial-clusters}
\end{figure}

Fig.~\ref{fig:aqi-spatial-clusters} shows a visualization of the learned clusters for the Air Quality dataset to complement the one provided in the paper~(Fig.~\ref{fig:air_clusters}). In particular, the figure shows each sensor's geographical location, the partitioning of the network into clusters, and a visualization of the pooled graph for the first level of the hierarchy.

\end{document}